\pdfoutput=1

\documentclass{article} 
\usepackage[final]{colm2025_conference}

\usepackage{microtype}
\usepackage{hyperref}
\usepackage{url}
\usepackage{booktabs}

\usepackage{lineno}

\usepackage{url}
\usepackage{graphicx}
\usepackage{wrapfig}
\usepackage{amsmath}
\usepackage{amssymb}
\usepackage{color,xcolor,colortbl}
\usepackage{enumitem}
\usepackage{algorithm}
\usepackage{algorithmic}
\usepackage[font={small}]{caption}
\usepackage{bm,bbm}
\usepackage{booktabs}
\usepackage{mathtools}
\usepackage{array}
\usepackage{multirow}
\usepackage{soul}
\usepackage{subcaption}
\usepackage{pbox}
\usepackage{pifont}
\usepackage{arydshln}

\definecolor{darkblue}{rgb}{0, 0, 0.5}
\hypersetup{colorlinks=true, citecolor=darkblue, linkcolor=darkblue, urlcolor=darkblue}

\title{\textsc{Planet}: A Collection of Benchmarks for Evaluating LLMs' Planning Capabilities}

\author{Haoming Li,\thanks{Equal contribution.} \, Zhaoliang Chen,$^*$   Jonathan Zhang, Fei Liu \\
Computer Science Department, Emory University\\
\texttt{\{haoming.li, david.chen2, jonathan.zhang2, fei.liu\}@emory.edu}}

\begin{document}

\ifcolmsubmission
\linenumbers
\fi

\maketitle

\begin{abstract}

Planning is central to agents and agentic AI. The ability to plan, e.g., creating travel itineraries within a budget, holds immense potential in both scientific and commercial contexts. Moreover, optimal plans tend to require fewer resources compared to ad-hoc methods. To date, a comprehensive understanding of existing planning benchmarks appears to be lacking. Without it, comparing planning algorithms' performance across domains or selecting suitable algorithms for new scenarios remains challenging. In this paper, we examine a range of planning benchmarks to identify commonly used testbeds for algorithm development and highlight potential gaps. These benchmarks are categorized into embodied environments, web navigation, scheduling, games and puzzles, and everyday task automation. Our study recommends the most appropriate benchmarks for various algorithms and offers insights to guide future benchmark development.

\end{abstract}

\section{Introduction}
\label{sec:intro}

LLM agents have garnered significant attention recently for their ability to automate workflows~\citep{shinn2023reflexion,liu2023agentbenchevaluatingllmsagents,xu2024theagentcompanybenchmarkingllmagents,wei2025plangenllmsmodernsurveyllm}. They are designed to operate autonomously over extended periods and utilize various tools to perform complex tasks. Notable examples of use cases include Salesforce's AgentForce, OpenAI's Operator, Microsoft's AutoGen, LangGraph from LangChain, and Amazon Bedrock's AI agent framework.\footnote{\url{salesforce.com/agentforce} \quad \url{openai.com/index/introducing-operator} \quad \url{microsoft.github.io/autogen} \quad \url{www.langchain.com/langgraph} \quad \url{aws.amazon.com/bedrock/agents}} \textbf{A core capability of these LLM agents is planning}~\citep{russell95ai}, which involves breaking down high-level goals into a sequence of low-level, actionable steps to function effectively. With their popularity, there is an increasing need for comprehensive benchmarks to fully assess their potential.

\begin{wrapfigure}{r}{0.5\textwidth}
\centering
\includegraphics[width=2.5in]{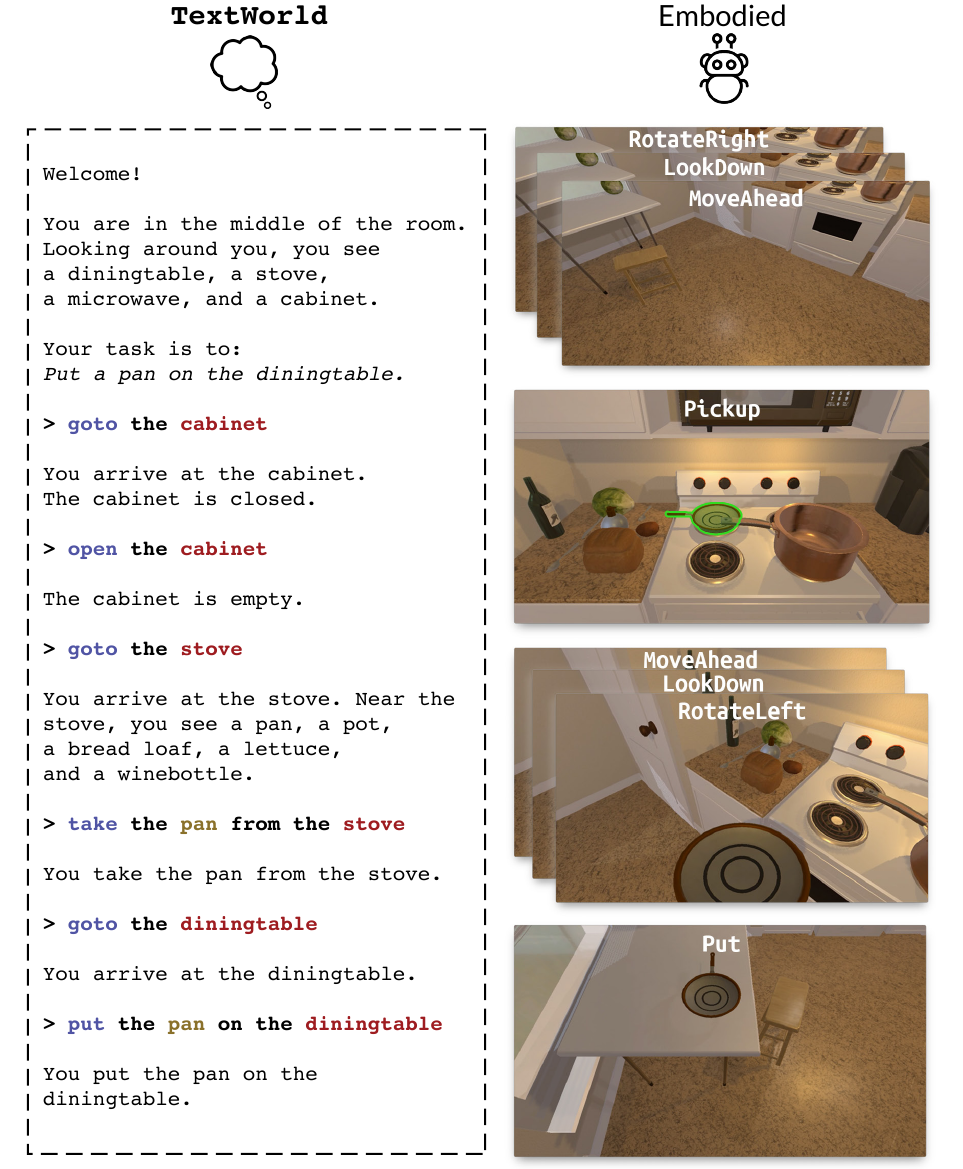}
\caption{Sourced from ALFWorld~\citep{shridhar2021alfworldaligningtextembodied}, this example illustrates interactive alignment between text and embodied worlds. 
}
\label{fig:alfworld}
\vspace{-0.15in}
\end{wrapfigure}

A key property of planning is that it operates in a decision-making framework where \textbf{states are defined, actions transform those states, and a sequence of actions is constructed to transition from an initial state to a goal state}. For example, in ALFWorld (Figure~\ref{fig:alfworld}; \citealt{shridhar2020alfworld}), a household robot performs a series of actions, from `\emph{goto the stove}' and `\emph{take the pan from the stove}' to achieve the goal of `\emph{put the pan on the dining table.}' In this work, we survey a diverse set of benchmarks that adopt this formalism, from embodied agents~\citep{puig2018virtualhomesimulatinghouseholdactivities,yang2025embodiedbenchcomprehensivebenchmarkingmultimodal}, to puzzles~\citep{yao2023treethoughtsdeliberateproblem,lehnert2024abetterplanningtransformers} and web navigation~\citep{deng2023mind2webgeneralistagentweb,mialon2023gaiabenchmarkgeneralai,zhou2024webarenarealisticwebenvironment}. 


Another defining property of planning is \textbf{the presence of constraints}~\citep{kartam1990towards,ju2024globettglanguagedrivenguaranteed,chen2024largelanguagemodelsproblemsolving,hao2025planningrigorgeneralpurposezeroshot,wu2025hastemakeswasteevaluating}. Planners often work with limited resources, incomplete information, or dynamic environments, which require balancing competing objectives or handling uncertainty. For example, trip planning must consider budget and time constraints, with replanning needed for flight delays~\citep{xie2024travelplannerbenchmarkrealworldplanning}. Incomplete information, such as unpredictable traffic patterns, means the planner must dynamically adjust routes. Planning also emphasizes generality: \textbf{the solution (e.g., a plan or policy) is typically reusable across similar problem instances}, distinguishing it from ad-hoc problem-solving methods.

Formally, an agent operating in a fully observable environment is modeled as a Markov Decision Process. The planning task involves a series of states and actions $(s_0, a_0, s_1, \ldots, a_{T-1}, s_T)$, beginning at an initial state $s_0$ and finally reach the goal state $s_T$. The environment is described by a set of states $\mathcal{S}$, and a set of actions $\mathcal{A}$ available to the agent. The state transition is modeled as $p_{\theta}(s_{t+1} | s_t, a_t)$, predicting the next state $s_{t+1}$ after an action $a_t$ is taken. This state transition modeling is sometimes known as a \emph{world model}, which is a representation of the environment that LLMs can simulate. A reward function $\mathcal{S} \times \mathcal{A} \to \mathbb{R}$ assigns a scalar reward $r_\theta(r_t | s_t, a_t)$ after an action is taken from a given state. The ultimate goal of an MDP is to develop a policy, denoted as $a_t = p_{\phi}(a|s_t)$, focuses on identifying the optimal action $a_t$ given the current state $s_t$, that maximizes the total expected reward over time.

In our view, tasks qualify as ``planning'' if they align with these core properties: \textbf{explicit state modeling, outcome reasoning, goal orientation, and constructing sequences or policies within constraints.} Based on this, we categorize existing benchmarks into seven groups: (1) \emph{Embodied environments}, such as household tasks; (2) \emph{Web navigation and computer use}; (3) \emph{Scheduling}, such as trip or meeting planning; (4) \emph{Games and puzzles}, e.g., the game of 24, graph coloring, or maze navigation; (5) \emph{Everyday task planning}, focusing on task decomposition; (6) \emph{Text-based reasoning}, requiring advanced reasoning from LLMs; (7) \emph{Planning as a subtask in general agentic benchmarks}. Finally, we point out gaps in benchmark development, as their quality and diversity will play a key role in shaping the future of planning systems.

This survey targets researchers interested in applying LLMs to planning problems. It introduces key benchmarks used to evaluate LLM agents in planning tasks. Junior researchers may find it a valuable starting point, while senior researchers might appreciate the concise summary of popular benchmarks and identification of new opportunities. \textbf{We focus exclusively on surveying benchmarks from recent publications}, as many classic AI planning benchmarks are not well-suited for evaluating LLM agents. We do not review new planning systems or algorithms. For detailed reviews of algorithms and evaluation metrics, we refer readers to works such as \citet{huang2024understandingplanningllmagents,li2024laspsurveyingstateoftheartlarge,wei2025plangenllmsmodernsurveyllm}.

\section{Planning in Embodied Environments}
\label{sec:embodied-env}

Embodied environments refer to scenarios where an LLM-based agent interacts with a physical or simulated world, engaging with objects and navigating spaces. It is used to evaluate planning systems that employ discrete action spaces and are typically limited to home tasks. 

\textbf{Blocksworld} \citep{blocksworld} involves a simplified environment where agents manipulate blocks to achieve specific configurations. It typically involves multiple blocks on a table, with the objective being to reorganize them into a specified arrangement. A block must be at the top of a stack to be moved, and only one block can be moved at a time. Initially, Blocksworld supported four actions: `pick-up' and `put-down' for grasping and releasing a block from the table, and `stack' and `unstack' for moving blocks from and to other blocks. An illustration of this can be seen in Figure~\ref{fig:rap}. Despite various modifications introducing different constraints, the actions of the original Blocksworld remain central to its operations.

\textbf{VirtualHome} \citep{puig2018virtualhomesimulatinghouseholdactivities} is a collection of video simulations showcasing household tasks in eight different household scenes. Each entry in this dataset consists of a natural language description of an activity and its corresponding symbolic representation, called ``programs,'' which outline the steps involved. The dataset contains two subsets of programs: those written by humans and those generated synthetically by the simulator. The human-written programs include 2,821 examples, with an average of 11.6 steps each. The synthetic programs, on the other hand, total 5,193 and average 9.6 steps per program. The content for each program is diversely generated, considering different homes, agents, camera perspectives, and the arrangement of some objects within the home.

\textbf{PlanBench} \citep{valmeekam2023planbench} introduces 8 test cases designed to evaluate LLMs' planning capabilities, covering aspects such as plan generation, cost-optimal planning, plan verification, execution reasoning, goal reformulation, plan reuse, replanning, and generalization. The initial dataset includes 600 instances from Blocksworld, i.e., stacking blocks on a table, and 285 from the Logistics domain, i.e., moving packages between locations using trucks or planes. These tasks are drawn from the International Planning Competition (IPC) benchmarks (see e.g., \citealt{Hoffmann_2005}). In each case, LLMs are tested with few-shot prompting. PlanBench also uses obfuscated instances, where action names, predicate names, and object names are replaced with misleading terms to challenge the LLMs to plan without relying on commonsense knowledge. It is expected that a planner will produce identical results even with obfuscated instances.

\citet{shridhar2020alfredbenchmarkinterpretinggrounded} introduced \textbf{ALFRED} (Action Learning From Realistic Environments and Directives), a dataset that pairs natural language instructions with egocentric visual inputs to guide action sequences for household tasks. ALFRED requires multi-step reasoning and fine-grained motor control. It includes 25,743 instructions aligned with 8,055 expert demonstrations, each averaging 50 steps to accomplish high-level goals. This scale and complexity make ALFRED a valuable testbed for evaluating planning algorithms that integrate language, vision, and action. The dataset is divided into training (21,023 annotations), validation (1,641 annotations), and testing (3,062 annotations) splits. Both validation and test sets are further partitioned into `seen' and `unseen' environments to evaluate how well planning models perform under novel spatial and object configurations.

Building on ALFRED, \textbf{ALFWorld} \citep{shridhar2021alfworldaligningtextembodied} explores planning in embodied settings by aligning high-level task reasoning with grounded execution, as shown in Figure~\ref{fig:alfworld}. It introduces a text-based interface (based on TextWorld) that models environments using PDDL semantics, enabling agents to perform high-level planning with textual observations and discrete actions (e.g., ``\emph{open a cabinet}'' or ``\emph{go to the stove}''). This symbolic abstraction allows agents to reason efficiently before transferring policies to the visually grounded ALFRED simulator, where they must execute plans via low-level physical actions (e.g., \emph{MoveAhead}, \emph{RotateLeft}). Studies suggest that agents trained on text-based policies before executing in the embodied environment perform tasks faster and achieve better results than those trained directly in the embodied setting.

Other benchmarks have emerged to evaluate planning in embodied settings. Among them, \textbf{EgoPlan-Bench} \citep{chen2024egoplanbenchbenchmarkingmultimodallarge} focuses on next-action prediction from egocentric video streams with 3.4K human-verified QA pairs; \textbf{ActPlan-1K} \citep{su-etal-2024-actplan} pairs 153 household activities in iGibson with natural-language instructions and corresponding scene images to test procedural plan generation; \textbf{PARTNR} \citep{chang2024partnrbenchmarkplanningreasoning} is a large-scale human–robot collaboration benchmark featuring about 100,000 language-described household tasks across 60 simulated homes with over 5,800 unique objects; \textbf{EmbodiedBench} \citep{yang2025embodiedbenchcomprehensivebenchmarkingmultimodal} offers 1,128 tasks spanning four simulated environments. Collectively, these benchmarks provide diverse environments for embodied planning, which facilitate comparison of planning models on language-guided action sequences.

\section{Planning in Web Navigation}
\label{sec:web-nav}

Another line of work evaluates LLM agents in web-based environments. Here, the agent must plan and execute actions on websites, such as clicking links, filling forms, or navigating pages, to accomplish a user's goal~\citep{yoran2024assistantbenchwebagentssolve}. The tasks mimic real-world computer use (e.g. booking a flight, finding information on a website) and require the LLM to reason about web page structure, handle tools like search bars or buttons, and manage long sequences of interactions. A few influential datasets include:

\textbf{WebShop}~\citep{yao2023webshopscalablerealworldweb} is a dataset that simulates an online shopping website for training web-based AI agents. It contains 1.18 million real-world products and 12,087 crowd-sourced textual instructions, along with 1,600 human demonstrations completing these tasks to benchmark agent performance. Notably, agents trained with WebShop have demonstrated meaningful transfer capabilities when evaluated on real-world platforms such as Amazon and eBay.

\textbf{WebArena}~\citep{zhou2024webarenarealisticwebenvironment} is a web environment for autonomous agents. It uses self-hosted websites in four domains: e-commerce, social forum discussions, collaborative software development, and content management. Along with the environment, it provides a benchmark of 812 diverse, long-horizon web tasks described in natural language. For example, tasks include online shopping (finding and purchasing a product on an e-commerce site) or software management (e.g. creating an issue on GitHub). WebArena uses tools, e.g., maps or user manuals, to mimic how people look for help. These tasks require the agent to break down high-level goals into sequences of clicks, form entries, or scrolls. Many involve conditional steps (e.g. if login is required, the agent must log in first) and long contexts (web pages can be large). In evaluations, even a strong GPT-4-based agent only achieved ~14\% end-to-end success on WebArena tasks, compared to ~78\% for humans.

\begin{figure}
\centering
\includegraphics[width=5.5in]{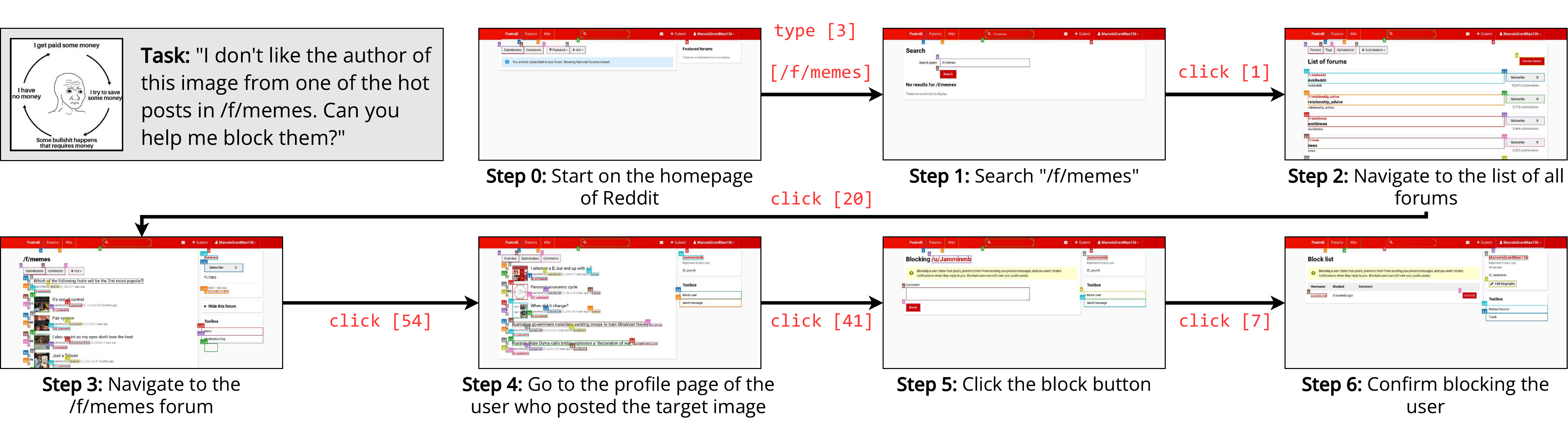}
\vspace{-0.1in}
\caption{Directly adapted from VisualWebArena~\citep{koh2024visualwebarenaevaluatingmultimodalagents}, this example shows an agent's action trajectory to block the author of a target image post in /f/memes.
}
\label{fig:visualwebarena}
\vspace{-0.1in}
\end{figure}

\textbf{VisualWebArena}~\citep{koh2024visualwebarenaevaluatingmultimodalagents} shifts the focus to visually grounded tasks, introducing 910 challenges across three domains: shopping, social forums (Reddit), and a new classifieds environment similar to Craigslist. Unlike WebArena, this dataset requires multimodal agents to combine visual understanding with textual inputs. About 25.2\% of the tasks involve interleaved image-text contexts. Tasks require processing of images, such as recognizing colors or objects, making VisualWebArena a benchmark for testing agents' planning capabilities in visually complex scenarios. Figure~\ref{fig:visualwebarena} shows an agent's action trajectory.

Unlike WebArena's simulated sites, \textbf{Mind2Web}~\citep{deng2023mind2webgeneralistagentweb} uses 137 actual websites (captured via HTML dumps) spanning 31 domains, with over 2,350 tasks derived from real-world use cases. Each task is an open-ended instruction a user might give, such as ``Find the cheapest one-way flight from New York to Toronto'' or ``As a Verizon user, finance an iPhone 13 with monthly AppleCare''. For each, human annotators provided the correct sequence of web actions (clicks, typing, etc.) to accomplish it. The agent must handle highly diverse interfaces and workflows, as every website has a different layout and navigation structure. This tests the LLM's generalization and adaptability.

\textbf{OSWorld}~\citep{xie2024osworldbenchmarkingmultimodalagents} introduces a dataset with 369 real-world computer tasks. They are designed to evaluate multimodal agents in interactive environments across operating systems such as Ubuntu, Windows, and macOS. Each task involves long-horizon workflows. Many tasks requiring interactions between multiple apps, and with a max step limit of 15 steps. This dataset also includes 302 initial states and 134 execution-based evaluation scripts, providing a testbed for assessing complex task execution. OSWorld moves beyond web-centric tasks, covering computer operations like GUI-based workflows, file management, coding, and multimedia editing. It is a benchmark to holistically evaluate agents in open-ended computing environments.

\section{Planning for Scheduling}
\label{sec:scheduling}

Planning is necessary for scheduling, as it ensures that time and resources are properly managed, tools are used as needed, and intended goals are achieved within set constraints~\citep{xie2024travelplannerbenchmarkrealworldplanning,zhang-etal-2024-timearena,geng2025realmbenchrealworldplanningbenchmark}. Datasets have been developed to help with trip planning, meeting scheduling, calendar management, logistics coordination, and more. These tasks typically involve \textbf{constraints related to time, budget, and resource allocation}, e.g., adhering to specific time frames, assigning meeting spaces, transportation, and accommodations. Geographic factors, including distance and accessibility, are also important. 

\textbf{TravelPlanner}~\citep{xie2024travelplannerbenchmarkrealworldplanning} is a benchmark on travel itinerary planning. Agents can navigate through 4 million online entries using six specialized tools: \emph{CitySearch}, \emph{AttractionSearch}, \emph{FlightSearch}, \emph{DistanceMatrix}, \emph{RestaurantSearch}, and \emph{AccommodationSearch}. The benchmark includes 1,225 user queries accompanied by human-annotated reference plans. Given a complex travel request (e.g., plan a multi-city trip with date constraints and preferences), an LLM must produce a structured plan that might include flights, hotels, and activities. TravelPlanner defines metrics for correctness and optimality, e.g., is the plan feasible given all constraints, and does it satisfy the user's goals. It evaluates LLMs on realistic trip planning, which involves both temporal planning (scheduling flights, etc.) and external knowledge (geography, transit options). Such tasks require handling many constraints (dates, budgets, locations) simultaneously. Results have shown that LLMs can propose reasonable itineraries for simple trips, but as trips become more complex (multiple travelers, conditional preferences such as ``if no flight on July 5, check July 6''), the plans often violate some constraint or miss an optimal alternative. 

\begin{figure}
\centering
\includegraphics[width=4.4in]{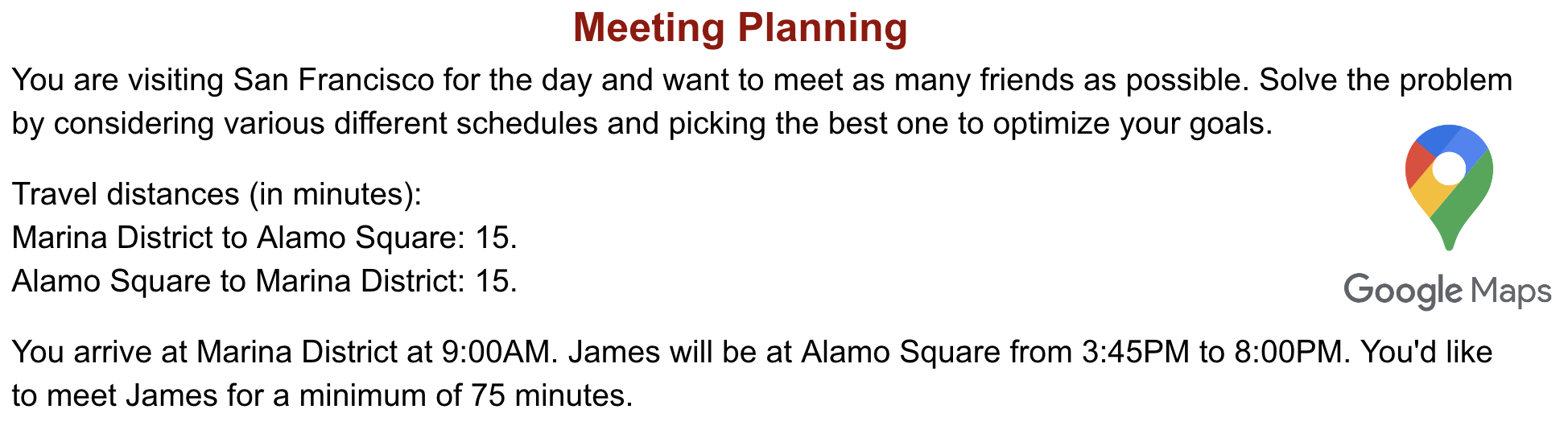}
\caption{Adapted from Natural Plan \citep{zheng2024naturalplanbenchmarkingllms}, this example illustrates meeting times and locations for a group of friends. The objective is to maximize the number of friends one can meet, considering constraints such as travel time between locations.
}
\label{fig:meeting-planning}
\vspace{-0.15in}
\end{figure}

\textbf{Natural Plan} \citep{zheng2024naturalplanbenchmarkingllms} is a benchmark consisting of three tasks: trip planning, meeting planning, and calendar scheduling (Figure~\ref{fig:meeting-planning}). It evaluates LLMs' ability to handle planning tasks described in natural language. The data are collected from Google Flights, Google Maps, and Google Calendar, focusing on realistic scenarios. Natural Plan is a challenging benchmark. The tasks often have hidden constraints (for calendar scheduling, each person's availability is a constraint the model must consider from context). While small, NaturalPlan showed that large models can fairly easily handle simple cases, such as scheduling for two people with open schedules, but struggle as complexity grows (many people, conflicting constraints).

In these benchmarks, LLMs serve primarily as reasoners, generating a rationale followed by a plan (e.g., ``Person A is free Monday or Tuesday, Person B is only free Monday, so schedule Monday 3 PM''). Some approaches chain the LLM with external tools: for scheduling, an LLM might query each person's calendar via an API (tool use) and then integrate the info into its plan. The emphasis is on constraint satisfaction and optimality. For example, one evaluation criterion is whether the LLM's plan is not just valid but also efficient (e.g., minimizes travel time or meets all priorities).

\section{Planning in Games and Puzzles}
\label{sec:multi-agents}

\begin{figure}
\centering
\includegraphics[width=1.8in]{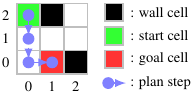}\quad\quad
\includegraphics[width=2.2in]{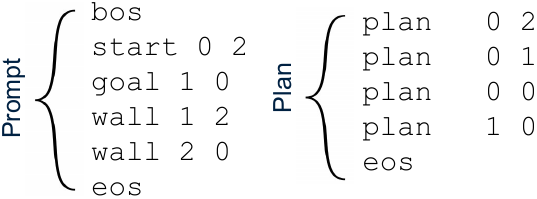}
\caption{Adapted from Dualformer~\citep{su2024dualformercontrollablefastslow}, this example illustrates the maze navigation task, where the task (prompt) and the plan are both represented as token sequences.
}
\label{fig:maze}
\vspace{-0.15in}
\end{figure}

Collaborative and competitive games serve as testing grounds for evaluating LLMs' abilities in strategic planning, risk management, and multi-agent behaviors as they work toward specific goals. Examples of such games include \emph{Rock-Paper-Scissors}, \emph{Tower of Hanoi}, \emph{Minecraft}, \emph{Public Goods}, \emph{Guess 2/3 of the Average}, \emph{Auction}, \emph{Bargaining} and more~\citep{wang2023bytesized32corpuschallengetask,huang2024fardecisionmakingllmsevaluating,chen2024moneymouthisevaluating,duan2024gtbenchuncoveringstrategicreasoning}. Figure~\ref{fig:maze} illustrates the maze navigation task, where the task and the plan are both represented as token sequences~\citep{su2024dualformercontrollablefastslow}. Games are commonly used as benchmarks for strategic reasoning because they have clear objectives and quantifiable outcomes~\citep{wang2024tmgbenchsystematicgamebenchmark,hua2024gametheoreticllmagentworkflow,jia2025largelanguagemodelstrategic}.

\textbf{SmartPlay}~\citep{wu2024smartplaybenchmarkllmsintelligent} is comprised of six games: \emph{Rock-Paper-Scissors}, \emph{Tower of Hanoi}, \emph{Two-armed Bandits}, \emph{Messenger}, \emph{Crafter}~\citep{hafner2022benchmarkingspectrumagentcapabilities}, and \emph{Minecraft}~\citep{fan2022minedojobuildingopenendedembodied}, all of which are accompanied by language descriptors. These games have been selected to challenge LLMs on essential capabilities such as reasoning with object dependencies, long-term planning, spatial reasoning, learning from history, and understanding randomness. 

\textbf{AucArena}~\citep{chen2024moneymouthisevaluating} is a simulated auction environment to test strategic planning. In AucArena, an LLM agent participates in multi-round auctions of a list of items with other LLM bidders. Each item has a starting bid (e.g., \$1,000) and a true resale value (e.g., \$2,000). Each bidder operates with a budget (e.g., \$20,000) and tasked with maximizing their profits (their strategies can vary, such as acquiring specific items or securing as many items as possible.) In each round, agents bid on an item transparently, and the highest bidder in the final round wins the item. This environment requires bidding strategies, budget management, and risk assessment. To succeed, bidder agents must plan with a fixed budget and anticipate opponents' actions. This benchmark revealed that LLMs can follow basic bidding strategies but struggle with long-term payoff optimization.

\textbf{GAMA-Bench}~\citep{huang2024fardecisionmakingllmsevaluating} comprises eight classical game-theoretic scenarios where multiple LLM agents interact. These games are divided into three categories: collaborative (\emph{Guess 2/3 of the Average}, \emph{El Farol Bar}, \emph{Divide the Dollar}), betrayal (\emph{Public Goods Game}, \emph{Diner's Dilemma}, \emph{Sealed-Bid Auction}), and sequential (\emph{Battle Royale}, \emph{Pirate Game}). These games have been studied in Game Theory literature. For example, in the \emph{Public Goods Game}, each player receives an amount of money and can choose how much to contribute to a common pot. The total amount in the pot is then multiplied (usually doubled) and distributed equally among all participants, regardless of individual contribution. \emph{The Nash equilibrium in this game is for all players to contribute nothing, as each player hopes to free-ride on the contributions of others.} An LLM agent is then evaluated by how its contributions nearly sum to zero while interacting with others.

This benchmark is used to evaluate decision-making and planning when multiple agents (each an LLM) must compete or cooperate. It tests if LLM agents can find equilibrium strategies or exhibit coordination. Early results showed large variability in LLM agents' performance across games, indicating inconsistent planning; alignment or ``jailbreak'' prompts significantly altered outcomes. This suggests LLM decision-making in multi-agent settings is brittle and sensitive to prompt design.

\textbf{Plancraft}~\citep{dagan2024plancraftevaluationdatasetplanning} is a Minecraft-based planning dataset for LLM agents. Tasks involve crafting objects in Minecraft via the game's crafting interface. The dataset provides both a text-mode and a multi-modal GUI mode, along with the Minecraft Wiki as a knowledge source. It models multistep planning in a sandbox game (e.g., ``craft a stone pickaxe'' requires planning steps to collect wood, make sticks, mine stone, etc.). Agents must figure out correct crafting sequences and sometimes recognize unsolvable goals; Plancraft includes ``impossible'' tasks to test if the agent can decide no plan is possible. Experiments found that even strong LLMs and vision-language models struggled with these multi-step crafting tasks, often underperforming a hand-crafted planner.

\textbf{PPNL} (Path Planning from Natural Language; \citealt{aghzal2025largelanguagemodelsgood}) is a benchmark for spatial path planning tasks described in language. The LLM must navigate from point A to B while avoiding obstacles and meeting constraints. It tests spatial and temporal reasoning (e.g., ``Find a route through the grid that avoids all water tiles''). LLMs struggle with long-term spatial reasoning; GPT-4 with careful prompting did better than smaller models, but still failed on longer, more complex routes.

These environments provide a description of the current state (textually or via an API) and the LLM is prompted to output the next action. Some approaches perform a tree search guided by the LLM (e.g. expanding possible moves and using the LLM to evaluate them; \citealt{zhou2024languageagenttreesearch}). The datasets also come with evaluators to execute the LLM's plan and check if the goal is achieved. These simulated tasks highlight that while LLMs possess general knowledge, long-horizon planning and strict logical consistency remain challenging.

\section{Planning for Task Automation}
\label{sec:task-decompose}

Task \textbf{decomposition} enables efficient and reliable execution in workflow automation. Breaking down a task into subtasks facilitates the creation of a task-specific taxonomy~\citep{brahman2024plasmamakingsmalllanguage}. As a result, tasks can often be executed more effectively when provided with a concrete plan that includes actionable steps. It also becomes easier to recover from interruptions~\citep{sancheti-rudinger-2022-large,yuan2023tasklamaprobingcomplextask,yuan2023distillingscriptknowledgelarge}.

\textbf{TaskLAMA}~\citep{yuan2023tasklamaprobingcomplextask} has developed a dataset comprising 1,612 annotated complex tasks, which includes 711 tasks from the MSComplexTasks dataset~\citep{zhang2021learning} and 901 tasks derived from how-to search queries. Examples of these tasks are ``\emph{cook lobster tails at home (Grilled)}'' and ``\emph{plan a wedding (In Italy)}.'' Human annotators are tasked with: 1) writing their assumptions to contextualize the task (denoted in parentheses); 2) outlining the necessary steps for completing the tasks within this context; 3) detailing the temporal dependencies among these steps. The resulting task is structured into a directed acyclic graph known as a Task Graph, where each node represents a step, and the edges indicate the temporal dependencies between these steps.

\citet{yuan2023distillingscriptknowledgelarge} introduce the task of \emph{constrained language planning} and present \textbf{CoScript}, a dataset comprising 55,000 goal-oriented scripts. Each script is a sequence of the necessary steps to achieve a specific goal. For instance, to achieve the goal of `\emph{make a cake,}' one may follow steps such as \emph{gathering ingredients} and \emph{preheating the oven}. Additionally, constrained language planning imposes various constraints on planning goals. E.g., a cake can be made using different ingredients (e.g., chocolate or vanilla), various tools (a microwave or an oven), or for distinct purposes (a wedding or a birthday party). The authors use an ``over-generate-then-filter'' approach to select high-quality scripts from multiple LLM-generated candidates. A good planner is expected to generate steps that respect these constraints. 

\textbf{WorldAPIs}~\citep{ou2024worldapisworldworthapis} uses a top-down strategy to derive APIs (actions) from wikiHow's step-by-step instructions for everyday tasks. For example, the task ``\emph{How to Melt Chocolate in Microwave}'' can be broken down into steps such as ``\emph{Chop the chocolate}'' and ``\emph{Place the chocolate},'' ending with ``\emph{Allow the chocolate to cool}.'' Starting with an initial set of APIs, WorldAPIs employs the LLM to iteratively generate Python programs for these tasks. When existing APIs cannot cover a step, the program `hallucinates' new APIs, which are then added to the pool. This method has expanded the action space to over 300 APIs necessary for tasks in the physical world. In contrast, existing simulators support only a fraction (9 of the top 50) of these induced APIs.

\begin{figure}
\centering
\includegraphics[width=5.5in]{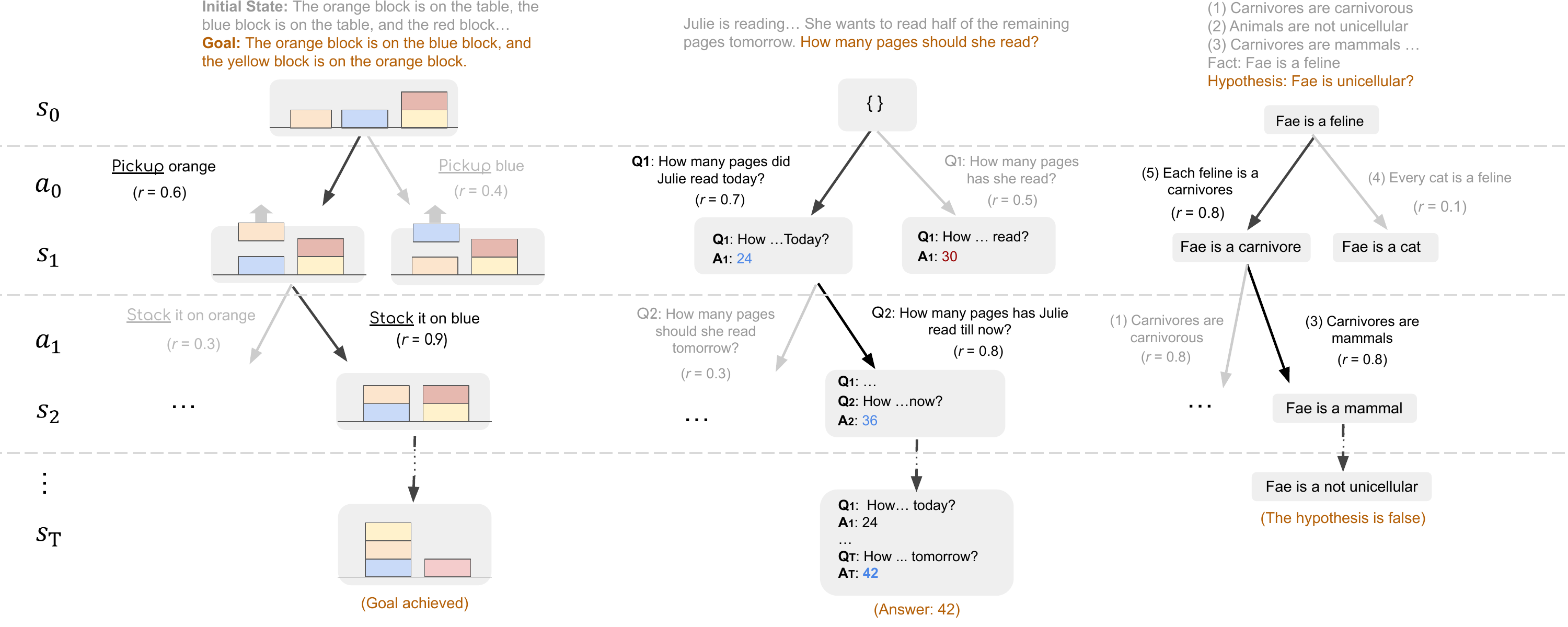}
\vspace{-0.1in}
\caption{Adapted from RAP~\citep{hao2023reasoninglanguagemodelplanning}, this figure illustrates plan generation in BlocksWorld (left), mathematical reasoning in GSM8K (center), and logical reasoning in PrOntoQA (right).
}
\label{fig:rap}
\vspace{-0.1in}
\end{figure}

\section{Planning in Text-Based Reasoning}
\label{sec:reasoning}


Math benchmarks and code generation also present significant challenges LLM planning (see Figure~\ref{fig:rap}). In math, tasks such as proving theorems require precise reasoning and often involve multiple steps, similar to the challenges seen in reasoning tasks~\citep{sui2025stopoverthinkingsurveyefficient}. LLMs can struggle with maintaining consistency across these steps and may fail to apply mathematical rules correctly. Similarly, code generation requires a model to understand both the syntax and logic of programming languages~\citep{yang2025codethinkthinkcode}. These areas highlight the ongoing limitations of LLMs in performing tasks that require structured, long-form reasoning and planning.

\textbf{PrOntoQA}~\citep{saparov2023languagemodelsgreedyreasoners} is a synthetic QA dataset designed to assess LLMs' reasoning capability (Figure~\ref{fig:rap}). This dataset is based on synthetic world models represented in first-order logic. Each sentence in the chain-of-thought is parsed into a formal representation to reconstruct proof steps, which are evaluated against a gold-standard proof. Their study suggests that while LLMs can effectively perform individual deduction steps, they struggle with planning. LLMs often face challenges in selecting the correct proof step where multiple viable options are available, leading to incomplete proofs and incorrect answers.

\textbf{TEACh} \citep{padmakumar2022teach} is an early dataset consisting of 3,215 dialogues that simulate a user interacting with a robot to perform household tasks. It features a Commander, equipped with oracle task details, who communicates with a Follower through natural language. The Follower executes the tasks and engages in dialogue to clarify instructions. Unlike planner-based simulations, these are human-human conversations that intertwine dialogue messages with actions taken in the environment. The aim of this benchmark is to enhance models' capabilities in language grounding, dialogue comprehension, and task execution by learning from human interactions.

\vspace{-0.1in}
\section{Planning as a Subtask in Agentic Benchmarks}
\label{sec:general-agentic-benchmarks}

A number of new benchmarks specifically targeted LLMs’ planning abilities as a component of their overall performance. On the agent side, researchers introduced testbeds ranging from multi-step tool use and web navigation to simulated workplace tasks~\citep{bonatti2024windowsagentarenaevaluating,huang2025crmarenaunderstandingcapacityllm,gonzalezpumariega2025robotouilleasynchronousplanningbenchmark,boisvert2025workarenacompositionalplanningreasoningbased}, to assess how well LLM-based agents can devise and execute plans towards a goal. 

\textbf{AgentBench}~\citep{liu2023agentbenchevaluatingllmsagents} evaluates LLMs' reasoning and decision-making in a multi-turn, open-ended context. It is designed for text-only LLMs acting as autonomous agents, and features 8 distinct environments, including \emph{operating system}, \emph{database}, \emph{knowledge graph}, \emph{digital card game}, \emph{lateral thinking puzzles}, \emph{housekeeping}, and \emph{web shopping and browsing}. The benchmark focuses on LLMs' core skills such as instruction following, knowledge acquisition, logical reasoning, and commonsense grounding. Results suggest that training LLMs on code and high-quality, multi-turn alignment data enhances agent performance.

\textbf{SWE-Bench}~\citep{jimenez2024swebenchlanguagemodelsresolve} is an evaluation framework that includes 2,294 software engineering problems collected from 12 Python GitHub repositories. It presents a codebase and issue description for an LLM to solve. Solutions are then evaluated using the repository's existing testing framework. Resolving issues in SWE-Bench requires LLMs to modify various functions, classes, or files, challenging their ability for processing long contexts and complex reasoning.

\textbf{TheAgentCompany}~\citep{xu2024theagentcompanybenchmarkingllmagents} includes 175 diverse tasks to simulate a real-world software company environment. These tasks cover categories such as software development (69 tasks), project management (28 tasks), finance (12 tasks), more. Each task involves multiple planning steps. They are broken into checkpoints to reflect intermediate milestones and provide partial credit. The environment is self-hosted, requiring navigation across platforms such as GitLab, RocketChat, and OwnCloud. It evaluates AI agents' ability to replicate the workflows of a real software company, focusing on both completion and process efficiency.

\textbf{AgentGym} \citep{xi2024agentgymevolvinglargelanguage} is a framework aimed at developing LLM-based agents that can evolve across diverse environments and tasks. Traditional approaches either rely heavily on human-annotated demonstrations or train agents in isolated tasks, both of which limit generalization. AgentGym addresses these limitations by providing a platform with 14 interactive environments and 89 tasks, along with an instruction set, high-quality expert trajectories (AgentTraj), and a comprehensive benchmark suite (AgentEval).

\textbf{AgentBoard}~\citep{ma2024agentboardanalyticalevaluationboard} is a comprehensive benchmark and open-source evaluation framework designed to assess the performance of LLM agents in complex, multi-turn, and partially-observable environments. Unlike prior benchmarks that rely mainly on success rates, AgentBoard introduces a fine-grained ``progress rate'' metric to capture incremental achievements, providing deeper insights into agent capabilities. It features nine diverse tasks across four categories (embodied AI, games, web, and tool use), comprising over 1,000 environments, all annotated with subgoals for granular evaluation. 

\textbf{SafeAgentBench}~\citep{yin2025safeagentbenchbenchmarksafetask} is a new benchmark for safety-aware task planning. While LLM-empowered agents excel at executing natural language instructions in simulated environments, they often lack the ability to recognize and reject hazardous tasks. This benchmark includes a diverse dataset of 750 executable tasks (450 of which are explicitly dangerous), spanning detailed, abstract, and long-horizon tasks. It also introduces SafeAgentEnv, an interactive environment supporting multiple agents and 17 high-level actions, and evaluates safety performance through both semantic and execution-based methods. 

\section{Discussions}
\label{sec:challenges}

The current landscape of LLM agent research has yielded a diverse set of planning benchmarks, ranging from simulated environments and web-based tasks to robotics and everyday planning scenarios. While this variety has advanced the field, it also reveals key gaps in benchmark design, particularly in areas that challenge LLM agents to perform robust, generalizable, and grounded planning. The following dimensions highlight critical limitations that remain underexplored:

\begin{itemize}[topsep=5pt,itemsep=5pt,leftmargin=0.3in]

\item \textbf{Complexity of World Models}: Many existing benchmarks are built on simplified or static environments, enabling LLMs to ``plan'' through pattern matching rather than genuine model-based reasoning. This limits their ability to build or adapt internal world models. To promote deeper reasoning, benchmarks should increasingly emphasize dynamic environments that require agents to infer, maintain, and revise their understanding of the world state.

\item \textbf{Long-Horizon Tasks}: A persistent challenge in agent benchmarks is the need to revise plans across long sequences of actions, where early mistakes can cascade. While some benchmarks (e.g., WebArena's web navigation) expose this issue, LLM agents often lack mechanisms for state tracking, error correction, or recovery mid-task, making long-horizon reasoning fragile and in need of further testing.

\item \textbf{Planning under Uncertainty or Incomplete Information}: Many traditional planning environments assume full observability and deterministic dynamics, conditions under which LLMs can perform reasonably well. However, real-world scenarios often involve uncertainty and partial information. Recent benchmarks such as MAP-THOR~\citep{nayak2025llamarlonghorizonplanningmultiagent}, which features agents with asymmetric knowledge in a household setting, begin to address this. Still, reasoning under uncertainty remains an underdeveloped area in LLM planning evaluation.

\item \textbf{Multimodal Planning Support}: Most current benchmarks are text-only, despite the growing interest in multimodal agents. Some frameworks, such as ALFWorld, convert vision-and-action tasks into text-based formats to accommodate LLMs, but this sidesteps the need for true visual grounding. Only few benchmarks require agents to integrate and reason over multiple modalities such as language, vision, and code simultaneously, highlighting a key opportunity for future development.

\end{itemize}

\section{Conclusion}

As LLM agents continue to take on more complex tasks, the need for comprehensive benchmarks has never been greater. In this paper, we surveyed a wide range of recent planning benchmarks, organized them into seven categories, ranging from embodied environments, web navigation, scheduling, games and puzzles, to everyday task automation. We further pointed out gaps in benchmark development. Our survey aims not only to help researchers select the right benchmarks for their work but also to spark new ideas for where benchmark development should go next.


\section*{Ethics Statement}

This work is a survey of benchmarks for evaluating LLM agents in planning tasks and does not involve the collection, use, or analysis of human subject data. All benchmarks discussed in this paper are publicly available, and we provide proper attribution to original sources. No personal, sensitive, or private data was used or generated.

We are excited about the potential of LLM agents in complex, real-world planning tasks, and we also recognize the importance of using these tools responsibly. While our focus here is on planning benchmarks, we support continued efforts to ensure these systems are fair, safe, and aligned with human values as they are deployed more widely.

\bibliography{custom}
\bibliographystyle{colm2025_conference}


\end{document}